\documentclass{cta-author}
\usepackage{amsmath,amssymb,amsfonts}
\usepackage{array}
\usepackage{booktabs}
\usepackage{multirow}
\usepackage{subfigure}
{}
{}
{}
\begin{document}
\bibliographystyle{unsrt}

\supertitle{Submission Template for IET Research Journal Papers}

\title{Considering Image Information and Self-similarity: A Compositional Denoising Network}

\author{\au{Jiahong Zhang$^{1,3}$}, \au{Yonggui Zhu$^{2\corr}$}, \au{Wenshu Yu$^{4}$}, \au{Jingning Ma$^{2}$}}

\address{\add{1}{State Key Laboratory of Media Convergence and Communication, Communication University of China, Beijing, China}
\add{2}{Data Science and Media Intelligence, Communication University of China, Beijing, China}
\add{3}{Neuroscience and Intelligent Media Institute, Communication University of China, Beijing, China}
\add{4}{School of Optoelectronic Science and Engineering, University of Electronic Science and Technology of China (UESTC), Chengdu, China}
\email{ygzhu@cuc.edu.cn}}

\begin{abstract}
Recently, convolutional neural networks (CNNs) have been widely used in image denoising. Existing methods benefited from residual learning and achieved high performance. Much research has been paid attention to optimizing the network architecture of CNN but ignored the limitations of residual learning. This paper suggests two limitations of it. One is that residual learning focuses on estimating noise, thus 
overlooking the image information. The other is that the image self-similarity is not effectively considered. This paper proposes a compositional denoising network (CDN), whose image information path (IIP) and noise estimation path (NEP) will solve the two problems, respectively. IIP is trained by an image-to-image way to extract image information. For NEP, it utilizes the image self-similarity from the perspective of training. This similarity-based training method constrains NEP to output a similar estimated noise distribution for different image patches with a specific kind of noise. Finally, image information and noise distribution information will be comprehensively considered for image denoising. Experiments show that CDN achieves state-of-the-art results in synthetic and real-world image denoising. Our code will be released on https://github.com/JiaHongZ/CDN.

\end{abstract}

\maketitle

\section{Introduction}
Image denoising is a commonly studied problem in the computer vision and shown to be important in medical image\cite{medical1,medical2}, remote-sensing image\cite{remote1}, mobile phone image\cite{dhdn}, etc. It aims to restore a corrupted image $x$ to the ground-truth clean image $y$, which can be modeled as $y=x-v$, where $v$ is the noise. Synthetic noisy images and real-world noisy images are studied in this paper. 

Recently, convolutional neural networks (CNNs) have been popularly adapted to image denoising. Zhang et al. \cite{dncnn} proposed DnCNN with residual learning and batch normalization to remove the additive white Gaussian noise (AWGN). Residual learning here is training networks to estimate the noise of the noisy image and then subtract it to get the corresponding clean image. Based on residual learning, CNNs obtained remarkable denoising results, completely exceeding the traditional methods such BM3D\cite{bm3d} and WNNM\cite{wnnm}. 
ResDNN \cite{resdnn} used ResNet-based blocks to deepen the network and achieved better denoising results than that of shallow networks. DNResNet \cite{DNResNet} and DRNet \cite{DRNet} then proposed more effective ResNet-based blocks to improve the network performance. However, deep CNNs suffer from gradient vanishing or exploding.  Some other methods constructed hierachical networks to alleviate this problem, such as two-path networks \cite{DudeNet,brdnet,adnet} and three-path U-Net-based networks \cite{MWCNN,dhdn,MCUNet}. To extracted effective features, attention-based networks also have received considerable attention \cite{NLRN,NHNet,PANNet,RID}.

Despite high performance in image denoising that the abovementioned methods achieved, the limitations of residual learning they used have not been well solved. Firstly, residual learning ignores the image information. The optimization target of residual learning can be expressed as:
\begin{equation}
    L(\theta) = L_{f}(x - f(x,\theta), y)
\end{equation}
, where $L_{f}$ is an arbitrary loss function, $\theta$ is trainable parameters and $f(\cdot)$ is the neural network. The right side of this equation is equivalent to $L_{f}(f(x,\theta), x-y)$, which actually gets the distance between estimated noise and actual noise of the noisy image $x$. That is, the residual learning is noise-to-noise. However, generating exquisite restored images also depends on the information of the image. This paper introduces an image information path (IIP), which is optimized by minimizing the Structure Similarity Index Measure (SSIM) \cite{ssim} loss between denoised images and ground-truth clean images. 
The proposed image-to-image training method makes IIP to extract the image information, solving the problem. 

Secondly, residual learning does not fully use the image self-similarity. However, the image self-similarity is essential in image restoration. Existing methods solved the problem by designing specific function modules such as non-local mechanism, which restores one pixel by using its neighbor pixels \cite{NCSR,NLRN,2015Nonlocal,nonlocal1,NHNet}. However, it costs high computation, especially when considering many neighbor pixels. We propose to utilize image self-similarity from the training perspective. Taking an image with the specific AWGN as an example, its noise distributions in different image patches exhibit to be similar. Therefore, we split the noisy image into patches and trained the proposed noise estimation path (NEP) to output similar noise estimations among these patches. Ablation experiments in Section \ref{ablation} show this similarity-based training method is effective. 

Based on the two points, we construct a compositional denoising network (CDN) with IIP, NEP, and an integration denoising module (IDM). IDM receives the outputs of IIP and NEP, generates the final estimated noise, and outputs a denoised image by residual subtraction. The main contributions of this paper are as follows:

(1) We suggest two limitations of the commonly used residual learning in image denoising, which have been overlooked in existing studies.

(2) We propose to utilize image information and image self-similarity to solve the limitations of residual learning. IIP with the image-to-image training and NEP with the similarity-based training in our work achieve these functions and are shown to be effective in the ablation experiments.

(3) Based on IIP and NEP, we propose a novel denoising network CDN, achieving state-of-the-art image denoising results on both synthetic and real-world datasets.

\begin{figure*}[!ht]
	\centerline{
		\includegraphics[width=1\textwidth]{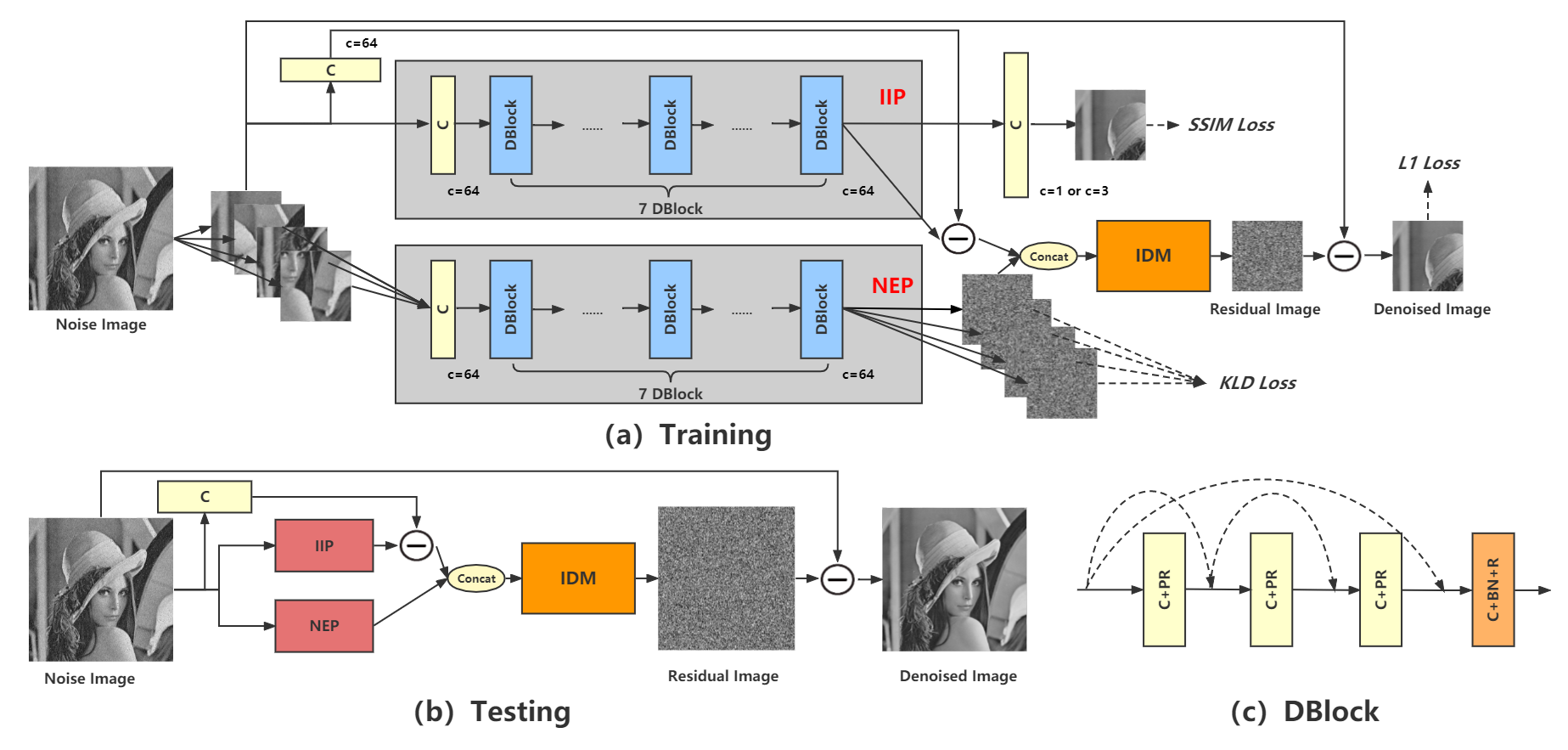}
	}
	\caption{The architecture of CDN. It contains three main modules, IIP, NEP, and IDM. (a) The training stage. A noisy image is split into several patches. The training target is removing the first patch's noise. Other patches are used in self-similarity training for NEP. IIP is optimized to extract image information in the image-to-image training method. NEP is trained by considering image self-similarity, which estimates the noise distribution. IDM integrates the output of IIP and NEP to get the final estimated noise. It is shown in Fig.~\ref{idm}. (b) shows the testing stage. It remove the noise of a complete image. (c) The basic CNN block of CDN, DBlock.}
	\label{cdn}
\end{figure*}

\section{Related Work}

\subsection{Residual learning for image denoising}
Residual learning was proposed in ResNet \cite{resnet} to solve the performance degradation problem with the increasing network depth. With such a learning strategy, the residual network will learn a residual mapping for a few stacked layers. Before ResNet, learning the residual mapping had already been adopted in some low-level vision tasks \cite{timofte2014a,kiku2013residual}. Zhang et al. \cite{dncnn} extended the residual learning to image denoising. Different from ResNet using many stacked residual units, they employed a single residual unit to predict the residual image. Recently, this residual learning has been commonly used in most deep denoising networks. Residual learning actually makes the network predict the noise of the image. However, achieving satisfactory denoising results also depends on the acquisition of image information. Furthermore, image self-similarity is also not considered in existing residual learning.

\subsection{Hierarchical networks for image denoising}
Deep neural networks have been widely applied to image denoising. Single path denoising networks such as DnCNN \cite{dncnn}, DNResNet\cite{DNResNet} and DRNet \cite{DRNet} achieved high denoising performance by using ever-deepening network structures. However, the increased depth makes models suffer from gradient vanishing or exploding. Hierarchical networks were proposed to use wide network structures to alleviate the problem. BRDNet \cite{brdnet} utilized two-path networks to increase the width of the network and thus obtained more features. DudeNet \cite{DudeNet} also contained two paths and further designed their different functions. For example, the top sub-path of DudeNet uses a sparse mechanism to extract global and local features. NHNet \cite{NHNet} used two sub-paths to process different resolutions of the noisy image. For high-resolution path, it employed a novel upsampling method with non-local mechanism to obtain effective features.  Some U-Net-based networks adopt a three-path structure to improve denoising performance. DHDN \cite{dhdn} replaced the convolution block in the original U-Net \cite{UNet} with dense blocks and obtained better denoising results. MCU-Net \cite{MCUNet} added an extra branch of atrous spatial pyramid pooling (ASPP) based on residual dense blocks. Sub-paths of these models extract different resolution image features, which will be fused at the end of the network for denoising. From a frequency domain perspective, MWCNN \cite{MWCNN} employed the multi-level wavelet packet transform (WPT) rather than upsampling and downsampling in U-Net-based methods and performed well on several image restoration tasks. 

This study uses a two-path hierarchical structure, and the sub-path functions are clearly defined. IIP extracts the image information, and NEP estimates the noise distribution using image self-similarity.

\begin{figure}[!ht]
	\centerline{
		\includegraphics[width=0.5\textwidth]{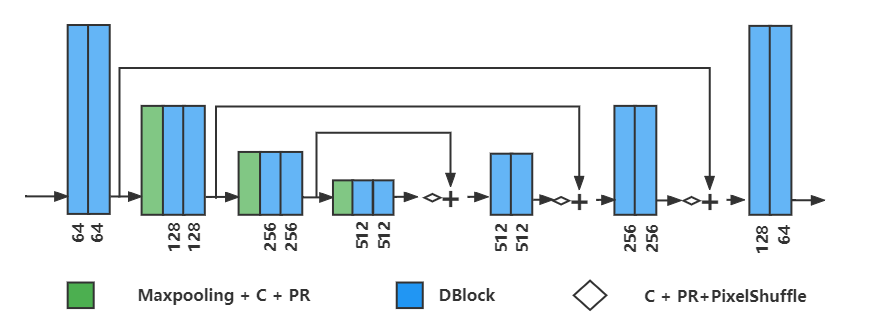}
	}
	\caption{The integration denoising module (IDM). It uses UNet-based archtecture and the number of output channels of each DBlock is shown at the bottom. DBlock is shown in Fig.~\ref{cdn} (c).}
	\label{idm}
\end{figure}

\section{The Proposed Method}

\subsection{Network Architecture}
The proposed CDN is shown in Fig.~\ref{cdn}, which consists of three main modules, IIP, NEP, and IDM. Here C denotes convolution layer, BN denotes batch normalization \cite{bn}, PR denotes parametric rectified linear unit \cite{prelu} and R denotes rectified linear unit \cite{relu}. During training, an input noisy image $x$ is divided into four patches $x_1,x_2,x_3,x_4$. This splitting operation aims to utilize the similarity of patches to train NEP. Here we empirically divide the image into four patches for the following reasons: 1) fewer patches do not take advantage of similarity; 2) more patches cause fewer noise samples per patches, thus can not sufficiently estimate the noise distribution. The training method of IIP is image-to-image. Without loss of generality, we choose the first patch $x_1$ and use $y_1$ to denote the ground-truth clean image of $x_1$. Finally, the denoised $x_1$ will be outputted by CDN, noted as $\Tilde{x_1}$.

When testing, the input of CDN is a complete noisy image, and the output is the denoised image. Convolution layers in CDN are set kernel size $(3 \times 3)$, stride 1, padding 1.

\begin{table*}[!t]
	\caption{PSNR (dB) results for different networks on Set12 with noise levels of 15, 25, 50.}
	\renewcommand\arraystretch{1.2}%\renewcommand\arraystretch{1.35}
	\begin{center}
		\begin{tabular}{|p{1.8cm}|c|c|c|c|c|c|c|c|c|c|c|c|c|}
			\hline
			\textbf{Images} & \textbf{\textit{C.man}}& \textbf{\textit{House}}& \textbf{\textit{Peppers}}& \textbf{\textit{Starfish}}& \textbf{\textit{Monarch}}& \textbf{\textit{Airplane}}& \textbf{\textit{Parrot}}& \textbf{\textit{Lena}}& \textbf{\textit{Barbara}}& \textbf{\textit{Boat}}& \textbf{\textit{Man}}& \textbf{\textit{Couple}}& \textbf{\textit{Average}} \\
			\hline
			\hline
			\textbf{Noise level}&\multicolumn{13}{|c|}{\textbf{$\sigma = 15$}} \\
			\hline
			\textbf{BM3D\cite{bm3d}} & \textbf{31.91}& \textbf{34.93}& \textbf{32.69}& \textbf{31.14}& \textbf{31.85}& \textbf{31.07}& \textbf{31.37}& \textbf{34.26}& \textbf{33.10}& \textbf{32.13}& \textbf{31.92}& \textbf{32.10}& \textbf{32.37} \\
			\hline	
			\textbf{DnCNN\cite{dncnn}} & \textbf{32.61}& \textbf{34.97}& \textbf{33.30}& \textbf{32.20}& \textbf{33.09}& \textbf{31.70}& \textbf{31.83}& \textbf{34.62}& \textbf{32.64}& \textbf{32.42}& \textbf{32.46}& \textbf{32.47}& \textbf{32.86} \\	
			\hline
			\textbf{FFDNet\cite{ffdnet}} & \textbf{32.43}& \textbf{35.07}& \textbf{33.25}& \textbf{31.99}& \textbf{32.66}& \textbf{31.57}& \textbf{31.81}& \textbf{34.62}& \textbf{32.54}& \textbf{32.38}& \textbf{32.41}& \textbf{32.46}& \textbf{32.77}\\	
			\hline
			\textbf{ResDNN\cite{resdnn}} & \textbf{32.73}& \textbf{34.99}& \textbf{33.23}& \textbf{32.11}& \textbf{33.20}& \textbf{31.65}& \textbf{31.87}& \textbf{34.57}& \textbf{32.56}& \textbf{32.39}& \textbf{32.42}& \textbf{32.43}& \textbf{32.85}\\	
			\hline
			\textbf{U-Net\cite{udnet}} & \textbf{32.33}& \textbf{34.79}& \textbf{33.16}& \textbf{32.00}& \textbf{32.94}& \textbf{31.64}& \textbf{31.84}& \textbf{34.46}& \textbf{32.43}& \textbf{32.30}& \textbf{32.34}& \textbf{32.31}& \textbf{32.71}\\	
			\hline
			\textbf{ADNet\cite{adnet}} & \textbf{\color{blue}32.81}& \textbf{35.22}& \textbf{\color{blue}33.49}& \textbf{32.17}& \textbf{33.17}& \textbf{31.86}& \textbf{31.96}& \textbf{34.71}& \textbf{32.80}& \textbf{32.57}& \textbf{32.47}& \textbf{32.58}& \textbf{32.98}\\
			\hline	
			\textbf{DudeNet\cite{DudeNet}} & \textbf{32.71}& \textbf{35.13}& \textbf{33.38}& \textbf{32.29}& \textbf{33.28}& \textbf{31.78}& \textbf{31.93}& \textbf{34.66}& \textbf{32.73}& \textbf{32.46}& \textbf{32.46}& \textbf{32.49}& \textbf{32.94}\\
			\hline
			\textbf{BRDNet\cite{brdnet}} & \textbf{32.80}& \textbf{35.27}& \textbf{33.47}& \textbf{32.24}& \textbf{33.35}& \textbf{31.82}& \textbf{32.00}& \textbf{34.75}& \textbf{32.93}& \textbf{32.55}& \textbf{32.50}& \textbf{32.62}& \textbf{33.03}\\	
			\hline				
			\textbf{\textit{NHNet\cite{NHNet}}} & \textbf{\color{red}32.95}& \textbf{\color{blue}35.40}& \textbf{\color{red}33.60}& \textbf{\color{blue}32.36}& \textbf{\color{blue}33.55}& \textbf{\color{red}31.98}& \textbf{\color{red}32.10}& \textbf{\color{blue}34.80}& \textbf{\color{blue}33.14}& \textbf{\color{blue}32.65}& \textbf{\color{blue}32.57}& \textbf{\color{blue}32.69}& \textbf{\color{blue}33.15}\\
			\hline
			\textbf{\textit{CDN}} & \textbf{32.73}& \textbf{\color{red}35.70}& \textbf{33.42}& \textbf{\color{red}32.40}& \textbf{\color{red}33.57}& \textbf{\color{blue}31.87}& \textbf{\color{blue}32.02}& \textbf{\color{red}34.91}& \textbf{\color{red}33.48}& \textbf{\color{red}32.73}& \textbf{\color{red}32.58}& \textbf{\color{red}32.73}& \textbf{\color{red}33.18}\\
			\hline
			\hline
			\textbf{Noise level}&\multicolumn{13}{|c|}{\textbf{$\sigma = 25$}} \\
			\hline
			\textbf{BM3D\cite{bm3d}} & \textbf{29.45}& \textbf{32.85}& \textbf{30.16}& \textbf{28.56}& \textbf{29.25}& \textbf{28.42}& \textbf{28.93}& \textbf{32.07}& \textbf{\color{blue}30.71}& \textbf{29.90}& \textbf{29.61}& \textbf{29.71}& \textbf{29.97} \\
			\hline
			\textbf{DnCNN\cite{dncnn}}& \textbf{30.18}& \textbf{33.06}& \textbf{30.87}& \textbf{29.41}& \textbf{30.28}& \textbf{29.13}& \textbf{29.43}& \textbf{32.44}& \textbf{30.00}& \textbf{30.21}& \textbf{30.10}& \textbf{30.12}& \textbf{30.43} \\
			\hline
			\textbf{FFDNet\cite{ffdnet}} & \textbf{30.10}& \textbf{33.28}& \textbf{30.93}& \textbf{29.32}& \textbf{30.08}& \textbf{29.04}& \textbf{29.44}& \textbf{32.57}& \textbf{30.01}& \textbf{30.25}& \textbf{30.11}& \textbf{30.20}& \textbf{30.44}\\	
			\hline
			\textbf{ResDNN\cite{resdnn}} & \textbf{30.17}& \textbf{32.99}& \textbf{30.73}& \textbf{29.24}& \textbf{30.30}& \textbf{29.00}& \textbf{29.38}& \textbf{32.31}& \textbf{29.70}& \textbf{30.11}& \textbf{30.04}& \textbf{29.96}& \textbf{30.33}\\	
			\hline
			\textbf{U-Net\cite{udnet}} & \textbf{30.18}& \textbf{33.18}& \textbf{30.91}& \textbf{29.38}& \textbf{30.41}& \textbf{29.18}& \textbf{29.57}& \textbf{32.59}& \textbf{30.19}& \textbf{30.25}& \textbf{30.10}& \textbf{30.14}& \textbf{30.51}\\	
			\hline
			\textbf{ADNet\cite{adnet}} & \textbf{30.34}& \textbf{33.41}& \textbf{31.14}& \textbf{29.41}& \textbf{30.39}& \textbf{29.17}& \textbf{29.49}& \textbf{32.61}& \textbf{30.25}& \textbf{30.37}& \textbf{30.08}& \textbf{30.24}& \textbf{30.58}\\	
			\hline
			\textbf{DudeNet\cite{DudeNet}} & \textbf{30.23}& \textbf{33.24}& \textbf{30.98}& \textbf{29.53}& \textbf{30.44}& \textbf{29.14}& \textbf{29.48}& \textbf{32.52}& \textbf{30.15}& \textbf{30.24}& \textbf{30.08}& \textbf{30.15}& \textbf{30.52}\\	
			\hline
			\textbf{BRDNet\cite{brdnet}} & \textbf{\color{red}31.39}& \textbf{33.41}& \textbf{31.04}& \textbf{29.46}& \textbf{30.50}& \textbf{29.20}& \textbf{29.55}& \textbf{32.65}& \textbf{30.34}& \textbf{30.33}& \textbf{30.14}& \textbf{30.28}& \textbf{30.61}\\	
			\hline		
			\textbf{\textit{NHNet\cite{NHNet}}} & \textbf{\color{blue}30.49}& \textbf{\color{blue}33.65}& \textbf{\color{red}31.20}& \textbf{\color{blue}29.72}& \textbf{\color{blue}30.68}& \textbf{\color{blue}29.34}& \textbf{\color{blue}29.65}& \textbf{\color{blue}32.76}& \textbf{30.70}& \textbf{\color{blue}30.44}& \textbf{\color{blue}30.20}& \textbf{\color{blue}30.40}& \textbf{\color{blue}30.77}\\
						\hline		
			\textbf{\textit{CDN}} & \textbf{30.45}& \textbf{\color{red}33.86}& \textbf{31.12}& \textbf{\color{red}29.78}& \textbf{\color{red}30.91}& \textbf{\color{red}29.29}& \textbf{\color{red}29.67}& \textbf{\color{red}32.95}& \textbf{\color{red}31.24}& \textbf{\color{red}30.60}& \textbf{\color{red}30.27}& \textbf{\color{red}30.52}& \textbf{\color{red}30.89}\\
			\hline			
			\hline
			\textbf{Noise level}&\multicolumn{13}{|c|}{\textbf{$\sigma = 50$}} \\
			\hline
			\textbf{BM3D\cite{bm3d}} & \textbf{26.13}& \textbf{29.69}& \textbf{26.68}& \textbf{25.04}& \textbf{25.82}& \textbf{25.10}& \textbf{25.90}& \textbf{29.05}& \textbf{\color{blue}27.22}& \textbf{26.78}& \textbf{26.81}& \textbf{26.46}& \textbf{26.72} \\
			\hline
			\textbf{DnCNN\cite{dncnn}} & \textbf{27.03}& \textbf{30.00}& \textbf{27.32}& \textbf{25.70}& \textbf{26.78}& \textbf{25.87}& \textbf{26.48}& \textbf{29.39}& \textbf{26.22}& \textbf{27.20}& \textbf{27.24}& \textbf{26.90}& \textbf{27.18} \\
			\hline
			\textbf{FFDNet\cite{ffdnet}} & \textbf{27.05}& \textbf{30.37}& \textbf{27.54}& \textbf{25.75}& \textbf{26.81}& \textbf{25.89}& \textbf{26.57}& \textbf{29.66}& \textbf{26.45}& \textbf{27.33}& \textbf{27.29}& \textbf{27.08}& \textbf{27.32} \\
			\hline		
			\textbf{ResDNN\cite{resdnn}} & \textbf{26.63}& \textbf{29.27}& \textbf{26.68}& \textbf{25.31}& \textbf{26.27}& \textbf{25.35}& \textbf{26.01}& \textbf{28.80}& \textbf{24.48}& \textbf{26.72}& \textbf{26.90}& \textbf{26.25}& \textbf{26.56} \\
			\hline
			\textbf{U-Net\cite{udnet}} & \textbf{27.42}& \textbf{30.48}& \textbf{27.67}& \textbf{25.92}& \textbf{26.94}& \textbf{25.89}& \textbf{26.66}& \textbf{\color{blue}29.84}& \textbf{27.02}& \textbf{27.42}& \textbf{27.30}& \textbf{27.17}& \textbf{27.48}\\	
			\hline
			\textbf{ADNet\cite{adnet}} & \textbf{27.31}& \textbf{30.59}& \textbf{27.69}& \textbf{25.70}& \textbf{26.90}& \textbf{25.88}& \textbf{26.56}& \textbf{29.59}& \textbf{26.64}& \textbf{27.35}& \textbf{27.17}& \textbf{27.07}& \textbf{27.37} \\
			\hline
			\textbf{DudeNet\cite{DudeNet}} & \textbf{27.22}& \textbf{30.27}& \textbf{27.51}& \textbf{25.88}& \textbf{26.93}& \textbf{25.88}& \textbf{26.50}& \textbf{29.45}& \textbf{26.49}& \textbf{27.26}& \textbf{27.19}& \textbf{26.97}& \textbf{27.30} \\
			\hline						
			\textbf{BRDNet\cite{brdnet}} & \textbf{27.44}& \textbf{30.53}& \textbf{27.67}& \textbf{25.77}& \textbf{26.97}& \textbf{25.93}& \textbf{26.66}& \textbf{29.73}& \textbf{26.85}& \textbf{27.38}& \textbf{27.27}& \textbf{27.17}& \textbf{27.45}\\	
			\hline						
			\textbf{\textit{NHNet\cite{NHNet}}} & \textbf{\color{blue}27.54}& \textbf{\color{blue}30.85}& \textbf{\color{red}27.84}& \textbf{\color{blue}26.24}& \textbf{\color{blue}27.10}& \textbf{\color{blue}26.00}& \textbf{\color{blue}26.76}& \textbf{29.83}& \textbf{27.19}& \textbf{\color{blue}27.46}& \textbf{\color{blue}27.32}& \textbf{\color{blue}27.28}& \textbf{\color{blue}27.62}\\	
			\hline						
			\textbf{\textit{CDN}} & \textbf{\color{red}27.70}& \textbf{\color{red}31.26}& \textbf{\color{blue}27.82}& \textbf{\color{red}26.29}& \textbf{\color{red}27.23}& \textbf{\color{red}26.06}& \textbf{\color{red}26.88}& \textbf{\color{red}30.07}& \textbf{\color{red}28.12}& \textbf{\color{red}27.65}& \textbf{\color{red}27.42}& \textbf{\color{red}27.52}& \textbf{\color{red}27.83}\\	
			\hline
		\end{tabular}
		\label{tabc}
	\end{center}
\end{table*}

\subsubsection{Image information path (IIP)}
IIP consists of one convolution layer and seven DBlocks, as shown in Fig.~\ref{cdn}. It is proposed to extract the image information for effectively estimating noise. During training, the input of IIP is $x_1$. IIP extracts the image features of $x_1$, and then the denoised image $x_{1_c}$ and noise estimation $x_{1_n}$ will be obtained based on image features. It is expressed as:
\begin{equation}
\begin{aligned}
    x_{1_c} &= Conv(IIP(x_1)) \\
    x_{1_n} &= Conv(x_1) - IIP(x_1)
\end{aligned}
\end{equation}
, where Conv is the convolution layer changing the number of feature channels. $ x_{1_c}$ is used to constrain IIP to extract image information by the image-to-image training method. Therefore, $y_1$ is the optimization target of $x_{1_c}$, where SSIM is chose as the loss function. For $x_{1_n}$, it is further processed in IDM. The SSIM loss is as follows:
\begin{equation}
L_{SSIM} = \frac{2\mu_{x_{1_c}}\mu_{y_1} + C_1}{\mu_{x_{1_c}}^2 + \mu_{y_1}^2 + C_1} \times \frac{2\sigma_{{x_{1_c}}{y_1}} + C_2}{\sigma_{x_{1_c}}^2 + \sigma_{y_1}^2 + C_2} 
\end{equation}
Here, $\mu$, $\sigma$ and $\sigma_{{x_{1_c}}{y_1}}$ denote the mean, standard deviation and covariance, respectively. $C1$ and $C2$ are the image-dependent constants, which provide stabilization against small denominators. 

For testing, IIP receives a complete image and outputs its noise based on the image information.

\subsubsection{Noise estimation path (NEP)}
The architecture of NEP is similar to IIP. During training, $x_1,x_2,x_3,x_4$ are inputted to NEP respectively and their estimated noise $n_1,n_2,n_3,n_4$ are outputted. We suggest that $n_1,n_2,n_3,n_4$ should have the similar distribution when noise in the input image is specific. Therefore, Kullback-Leibler divergence (KLD) is used to evaluate the distance of these noise distributions:
\begin{equation}
D_{KL}(P \lVert Q)  = \sum P(x)\log\frac{P(x)}{Q(x)}
\end{equation}
, where P and Q are probability distributions. The sum of these distances forms the loss function:
\begin{equation}
L_{KLD} = \sum_{i=1}^4\sum_{j=1,j\neq i}^4 D_{KL}(n_i \lVert n_j)
\end{equation}
By minimizing $L_{KLD}$, this similarity-based training method solves the limitation of residual learning. 

For testing, NEP receives a complete noisy image and outputs its noise distribution.

\subsubsection{Integration denoising module (IDM)}
IDM is proposed to integrate the outputs from IIP and NEP. It is a U-Net-based network shown in Fig.~\ref{idm}. DBlock is used as the basic feature extraction module, and PixelShuffle is the upsampling method based on the efficient sub-pixel convolution \cite{pixelshuffle}. IDM outputs the final estimated noise and then gets the denoised image $\Tilde{x_1}$ by residual subtraction. We use L1 loss as the loss function of $\Tilde{x_1}$ and ground-truth clean image $y_1$:
\begin{equation}
L1 = \frac{1}{N} \rvert {\Tilde{x_1} - y_1} \rvert
\label{q1}
\end{equation}

Totally, the training loss $L$ contains $L_{SSIM}$, $L_{KLD}$ and $L1$:
\begin{equation}
L = L_{SSIM} + L_{KLD} + L1
\end{equation}

\subsection{Training setting}
CDN is implemented by pytorch 1.5.1 based on python 3.5 and cuda 9.2. Experiments run on NVIDIA Tesla P100 GPUs. We use Adam \cite{adam} algorithm with initial learning rate of 0.0002, weight decay of 0.0001 and a mini-batch size of 64 to optimize the trainable parameters. The learning rate will decrease with the increment of training epochs. Data augmentations are adopted to network training, which randomly splits the images into 128x128 patches and flips them horizontally and vertically.

% \begin{figure}[ht]
% 	\centering{\includegraphics[width=0.5\textwidth]{gray2.png}}
% 	\caption{Denoising results of image Monarch from Set12 with noise level 50 using different methods: (a) noisy image/14.71 dB, (b) BRDNet/26.97 dB, (c) NHNet/27.10 dB, (d) CDN/27.23 dB.}
% 	\label{gray_contrast1}
% \end{figure}

\begin{table*}[!t]
	\caption{Results for different networks on BSD68}
	\renewcommand\arraystretch{1.1}%\renewcommand\arraystretch{1.5}
	\begin{center}
		\begin{tabular}{|c|c|c|c|c|c|c|c|c|c|c|c|c|}
			\hline
			\textbf{Network} & \textbf{\textit{BM3D\cite{bm3d}}}&  \textbf{\textit{DnCNN\cite{dncnn}}}& \textbf{\textit{FFDNet\cite{ffdnet}}}& \textbf{\textit{ADNet\cite{adnet}}}& \textbf{\textit{DudeNet\cite{DudeNet}}}& \textbf{\textit{BRDNet\cite{brdnet}}}&
			\textbf{\textit{U-Net\cite{udnet}}}&
			\textbf{\textit{RIDNet\cite{00028}}}& \textbf{\textit{NHNet}} & \textbf{\textit{CDN}} \\			
			\hline
			\textbf{$\sigma=15$}&\textbf{31.07}&\textbf{31.72}&\textbf{31.62}&\textbf{31.74}&\textbf{31.78}&\textbf{31.79}& \textbf{31.54} &\textbf{31.81}&\textbf{\color{blue}31.85} & \textbf{\color{red}31.89}\\
			\hline
			\textbf{$\sigma=25$}&\textbf{28.57}&\textbf{29.23}&\textbf{29.19}&\textbf{29.25}&\textbf{29.29}&\textbf{29.29}& \textbf{29.13}&\textbf{29.34} &\textbf{\color{blue}29.37}& \textbf{\color{red}29.44}\\
			\hline
			\textbf{$\sigma=50$}&\textbf{25.62}&\textbf{26.23}&\textbf{26.30}&\textbf{26.29}&\textbf{26.31}&\textbf{26.36}& \textbf{26.39}&\textbf{\color{blue}26.40} &\textbf{\color{red}26.43}&\textbf{26.39}\\
			\hline
		\end{tabular}
		\label{tabc2}
	\end{center}
\end{table*}

\section{Experiments}

\subsection{Datasets}

\subsubsection{Synthetic noise datasets}
\begin{figure}[!ht]
	\centerline{
		\includegraphics[width=0.5\textwidth]{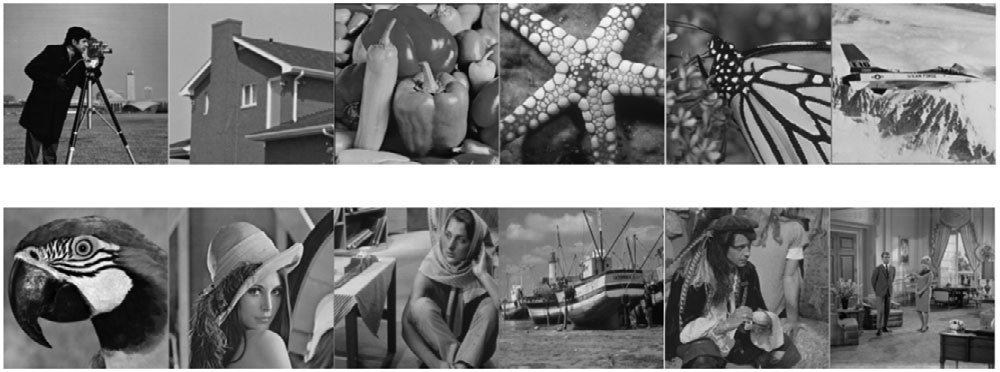}
	}
	\caption{Images in Set12, which are C.man, House, Peppers, Starfish,
Monarch, Airplane, Parrot, Lena, Barbara, Boat, Man and Couple in order}
	\label{set12}
\end{figure}

DIV2K \cite{DAVIS} is commonly used in image processing, containing 800 images for training, 100 for validation, and 100 for testing. We use the training set of DIV2K to train CDN. For testing, we use gray-scale image datasets Set12 and BSD68 \cite{set12} and color-scale image datasets Set5 \cite{set5} and Kodak24 \cite{kodak24}. Fig.~\ref{set12} shows images of Set12, which contains C.man, House, Peppers, etc. Images in these datasets are all clean and their corresponding synthetic noisy images are generated by adding the AWGN. We refers to the AWGN generation algorithm of \cite{dncnn}, in which the noise level is determined by the standard deviation $\sigma$. 
Three noise levels, $\sigma=15$, $\sigma=25$, and $\sigma=50$ are chose to trained and tested CDN.

\begin{figure*}[ht]
	\centering{\includegraphics[width=1\textwidth]{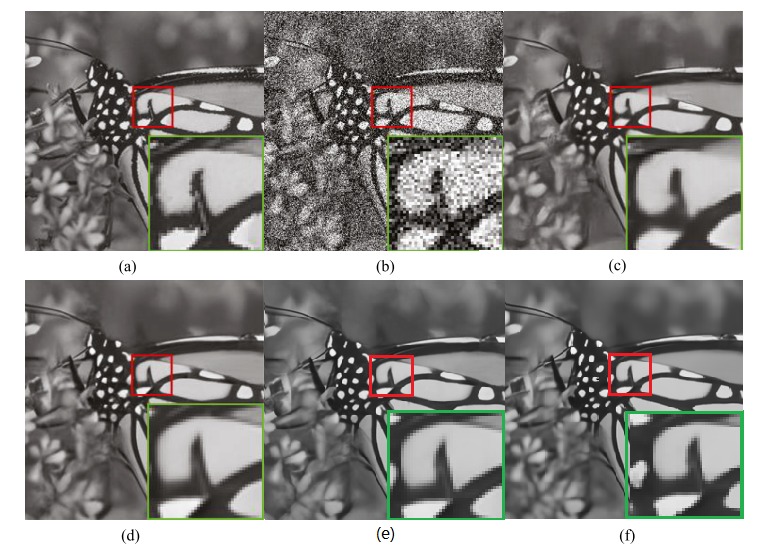}}
	\caption{Denoising results of the image Monarch from Set12 with noise level $\sigma=50$: (a) original image, (b) noisy image/14.71 dB, (c)  DnCNN\cite{dncnn}/26.78 dB, (d) BRDNet\cite{brdnet}/26.97 dB, (e) MHCNN\cite{mhcnn}/27.12 dB, and (f) CDN/27.21 dB. }
	\label{gray_contrast}
\end{figure*}

\begin{figure*}[ht]
	\centering{\includegraphics[width=1\textwidth]{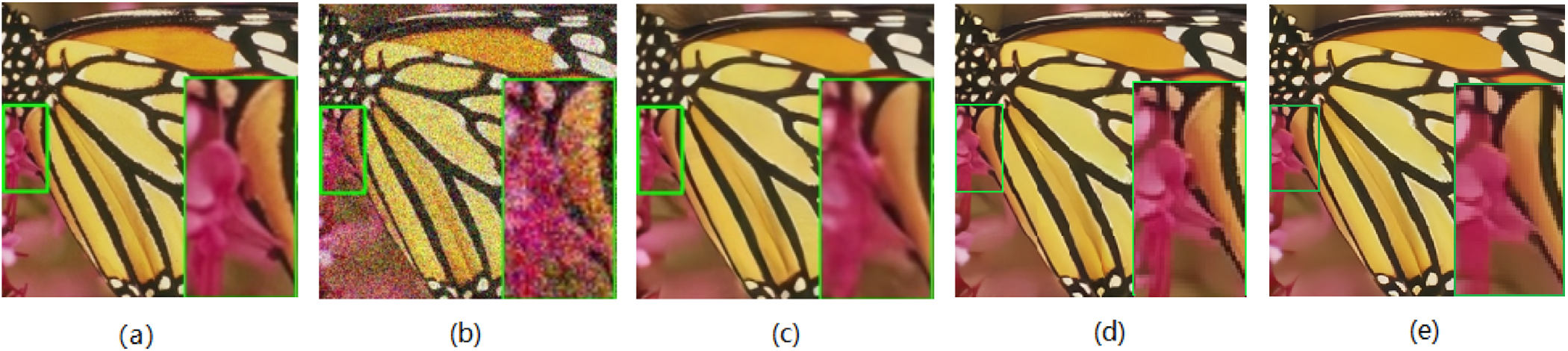}}
	\caption{Denoising result on butterfly from Set5 at noise level $\sigma=50$. (a) clear image, (b)noisy image, (c) VDN, (d) NHNet, (e) CDN. }
	\label{color_contrast}
\end{figure*}
\begin{figure*}[ht]
	\centering{\includegraphics[width=1\textwidth]{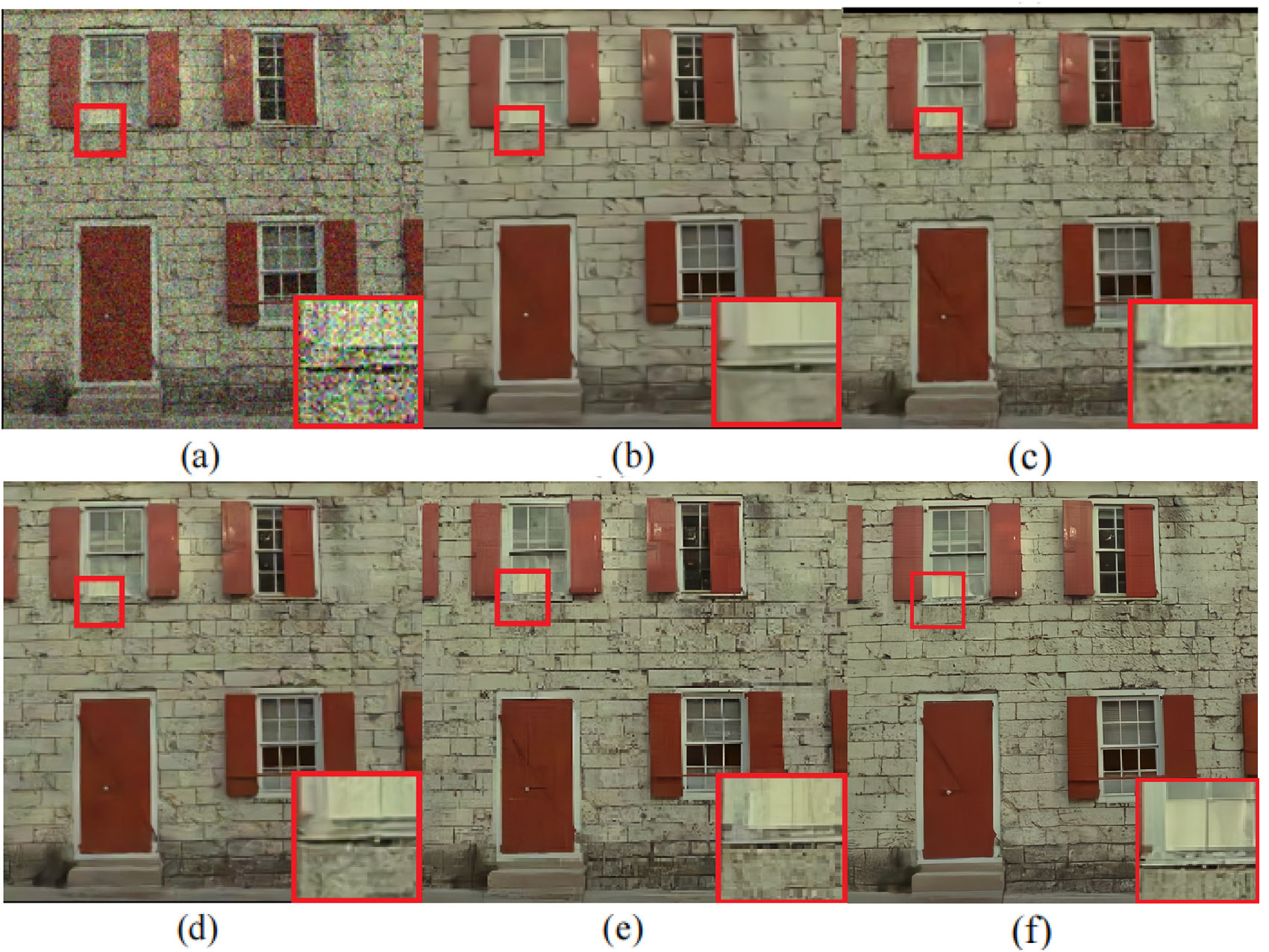}}
	\caption{Denoising result on the department wall from Kodak24 at noise level $\sigma=50$.  (a) noisy image, (b) DnCNN/25.80 dB, (c) BRDNet/26.33 dB, (d) FFDNet/26.13 dB, (e) NHNet/26.49 dB, and (f) CDN/29.53 dB. }
	\label{color_contrast2}
\end{figure*}

\begin{table}[t]
	\caption{Color image denoising results of different networks}
	\renewcommand\arraystretch{1.3} %\renewcommand\arraystretch{1.5}
	\begin{center}
		\begin{tabular}{|p{1.5cm}|c|c|c|c|c|}
			\hline
			\textbf{Datasets} & \textbf{\textit{Methods}}& \textbf{\textit{$\sigma=15$}}& \textbf{\textit{$\sigma=25$}}& \textbf{\textit{$\sigma=50$}} \\			
			\hline
			\multirow{5}{*}{Set5} &\textbf{\textit{CBM3D\cite{bm3d}}} & \textbf{33.42} & \textbf{30.92} & \textbf{28.16}\\
			& \textbf{\textit{FFDNet\cite{ffdnet}}} & \textbf{34.30} & \textbf{32.10} & \textbf{29.25}\\
			& \textbf{\textit{VDN\cite{vdn}}} & \textbf{34.34} & \textbf{32.24} & \textbf{29.47}\\						
			& \textbf{\textit{NHNet\cite{NHNet}}} & \textbf{\color{red}34.80} & \textbf{\color{blue}32.56} & \textbf{\color{blue}29.64}\\
			& \textbf{\textit{CDN}} & \textbf{\color{blue}34.70} & \textbf{\color{red}32.58} & \textbf{\color{red}29.66}\\
			\hline
			\multirow{9}{*}{Kodak24} &\textbf{\textit{CBM3D\cite{bm3d}}} & \textbf{34.28} & \textbf{31.68} & \textbf{28.46}\\
			& \textbf{\textit{FFDNet\cite{ffdnet}}} & \textbf{34.55} & \textbf{32.11} & \textbf{28.99}\\
			& \textbf{\textit{DnCNN\cite{dncnn}}} & \textbf{34.73} & \textbf{32.23} & \textbf{29.02}\\
			& \textbf{\textit{ADNet\cite{adnet}}} & \textbf{34.76} & \textbf{32.26} & \textbf{29.10}\\
			& \textbf{\textit{DudeNet\cite{DudeNet}}} & \textbf{34.81} & \textbf{32.26} & \textbf{29.10}\\
			& \textbf{\textit{BRDNet\cite{brdnet}}} & \textbf{34.88} & \textbf{32.41} & \textbf{29.22}\\
			& \textbf{\textit{NHNet\cite{NHNet}}} & \textbf{\color{blue}35.02} & \textbf{\color{blue}32.54} & \textbf{\color{blue}29.41}\\
			& \textbf{\textit{CDN}} & \textbf{\color{red}35.05} & \textbf{\color{red}32.57} & \textbf{\color{red}29.54}\\
			\hline
		\end{tabular}
		\label{tabcolor}
	\end{center}
\end{table}

\begin{figure*}[!ht]
	\centerline{
		\includegraphics[width=1\textwidth]{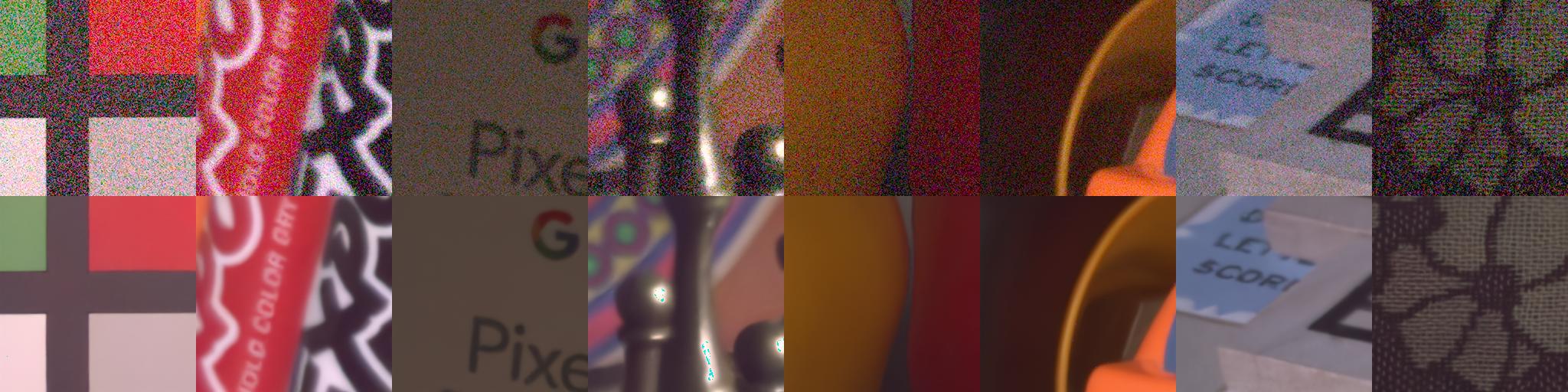}
	}
	\caption{Denoising results of CDN on SIDD. The noisy images are arranged at the top, and the denoised images are at the bottom.}
	\label{figreal}
\end{figure*}

\begin{table*}[ht]
	\renewcommand\arraystretch{1.1}
	\centering
	\caption{Denoising results of different networks on real-world noise datasets.}
	\begin{tabular}{p{2cm}|p{2cm}p{2cm}p{2cm}p{2cm}p{2cm}p{2cm}p{2cm}p{2cm}p{3cm}}
		\hline
		Test Data & \multicolumn{8}{c}{\textbf{ SIDD validation}}              \\ \hline
		Method     &
		\multicolumn{1}{c|}{BM3D\cite{bm3d}} &
		\multicolumn{1}{c|}{WNNM\cite{wnnm}} &\multicolumn{1}{c|}{CBDNet\cite{cbdnet}}    & \multicolumn{1}{c|}{RIDNet\cite{RID}}    & \multicolumn{1}{c|}{VDN\cite{vdn}}    & \multicolumn{1}{c|}{MHCNN\cite{mhcnn}}  &
		\multicolumn{2}{c}{CDN}  \\ \hline
		PSNR & \multicolumn{1}{c|}{ 25.65} & \multicolumn{1}{c|}{  25.78}
		& \multicolumn{1}{c|}{ 38.68} & \multicolumn{1}{c|}{38.71} & \multicolumn{1}{c|}{39.28} & \multicolumn{1}{c|}{\textbf{\color{blue}39.06}} & \multicolumn{2}{c}{\textbf{\color{red}39.36}}\\ 
		SSIM  & \multicolumn{1}{c|}{ 0.685}
		& \multicolumn{1}{c|}{ 0.685}
		& \multicolumn{1}{c|}{ 0.809} & \multicolumn{1}{c|}{\textbf{\color{blue}0.914}} & \multicolumn{1}{c|}{0.909} & \multicolumn{1}{c|}{\textbf{\color{blue}0.914}}& \multicolumn{2}{c}{\textbf{\color{red}0.918}}\\ \hline
		\hline
		Test Data & \multicolumn{8}{c}{\textbf{DND}}  \\
		\hline
		Method &\multicolumn{1}{c|}{BM3D\cite{bm3d}} &
		\multicolumn{1}{c|}{WNNM\cite{wnnm}}     &\multicolumn{1}{c|}{CBDNet\cite{cbdnet}}    & \multicolumn{1}{c|}{RIDNet\cite{RID}}    & \multicolumn{1}{c|}{VDN\cite{vdn}}    & \multicolumn{1}{c|}{PAN-Net\cite{PANNet}} & \multicolumn{1}{c|}{MHCNN\cite{mhcnn}} & 
		\multicolumn{1}{c}{CDN}\\
		\hline
		PSNR & \multicolumn{1}{c|}{34.51}
		& \multicolumn{1}{c|}{34.67}
		& \multicolumn{1}{c|}{38.06} & \multicolumn{1}{c|}{39.26} & \multicolumn{1}{c|}{39.38} & \multicolumn{1}{c|}{\textbf{\color{blue}39.44}} & \multicolumn{1}{c|}{\textbf{\color{red}39.52}} & \multicolumn{1}{c}{\textbf{\color{blue}39.44}}\\

		SSIM   & \multicolumn{1}{c|}{0.851} 
		& \multicolumn{1}{c|}{ 0.865} 
		& \multicolumn{1}{c|}{0.942} & \multicolumn{1}{c|}{\textbf{\color{red}0.953}} & \multicolumn{1}{c|}{\textbf{\color{blue}0.952}} & \multicolumn{1}{c|}{\textbf{\color{blue}0.952}} & \multicolumn{1}{c|}{0.951} & \multicolumn{1}{c}{0.951} \\
		\hline
	\end{tabular}
	\label{real}
\end{table*}

\begin{table*}\footnotesize
	\caption{Results (PSNR/SSIM) of ablation experiments on Set12.}
	\renewcommand\arraystretch{1.1}
	\begin{center}
		\begin{tabular}{c|p{1.5cm}|p{1.5cm}|p{1.5cm}|p{1.5cm}|p{1.5cm}|p{1.5cm}}
			\hline
			\textbf{Methods}& \multicolumn{2}{c|}{\textbf{\textit{$\sigma=15$}}}& \multicolumn{2}{c|}{\textbf{\textit{$\sigma=25$}}}& \multicolumn{2}{c}{\textbf{\textit{$\sigma=50$}}} \\
			\hline	
			&  \textit{PSNR} &  \textit{SSIM} & \textit{PSNR} &  \textit{SSIM} & \textit{PSNR} &  \textit{SSIM} \\
			\hline	
			\textit{CDN-IIP(R)} & 33.14 & 0.9080 & 30.82 & 0.8712 & 27.77 & 0.8050\\
			\textit{CDN-NEP(R)} & 33.02 & 0.9058 & 30.57 & 0.8659 & 27.54 & 0.7991\\
			\textit{CDN-SSIM} & 33.10 & 0.9075 & 30.80 & 0.8709 & 27.76 & 0.8052\\
			\textit{CDN-KLD} & 33.15 & 0.9083 & 30.83 & 0.8709 & 27.78 & 0.8051\\
			\textit{CDN-SSIM-KLD} &  33.05 & 0.9069 & 30.78 & 0.8703 & 27.75 & 0.8049\\
			\textit{CDN} & \textbf{33.18} & \textbf{0.9089} & \textbf{30.89} & \textbf{0.8724} & \textbf{27.83} & \textbf{0.8074}\\
			\hline
		\end{tabular}
		\label{contrast1}
	\end{center}
\end{table*}

\begin{figure*}[!ht]
	\centerline{
		\includegraphics[width=1\textwidth]{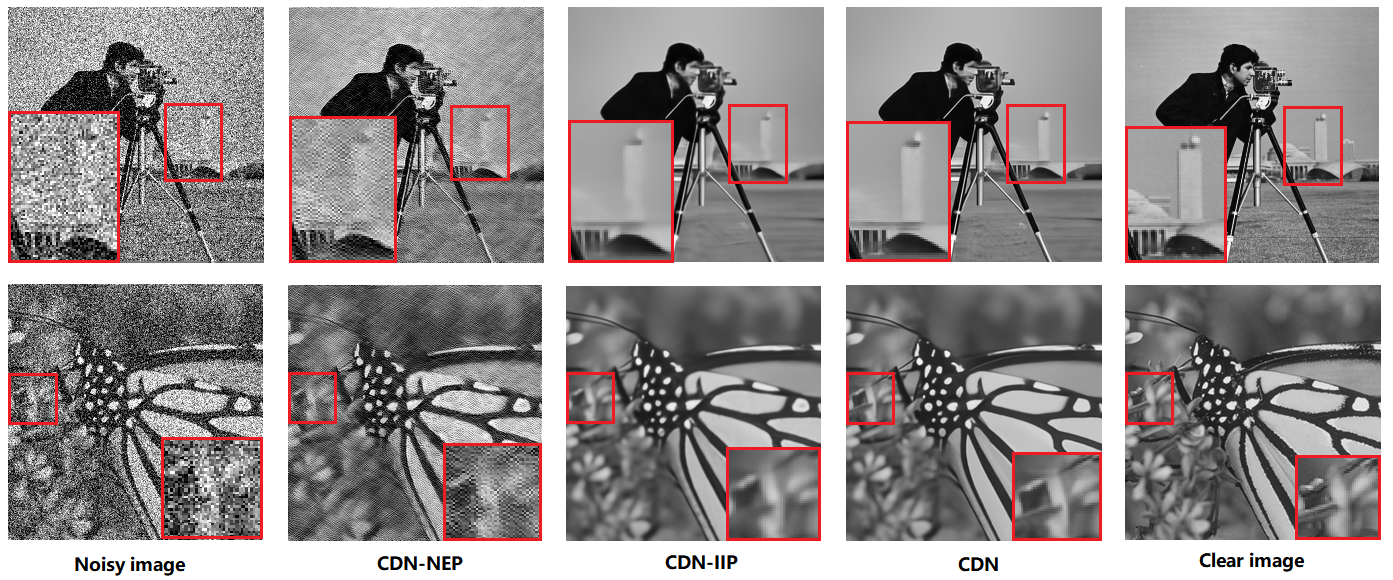}
	}
	\caption{Visual denoising results comparision of the ablation models of CDN. Images are from Set12. The left and right columns show the images with noise level $\sigma=50$ and the clear images, respectively. The middle three columns list the denoised images from CDN and its ablation variants. }
	\label{comp}
\end{figure*}

\subsubsection{Real-world noise datasets}
Real-world noise images are directly obtained in natural environment. Here, we use the training set of the Smartphone Image Denoising DATA (SIDD) sRGB track \cite{sidd} to train CDN. It contains 160 scene instances captured by five smartphone cameras under different lighting conditions and camera settings. There are two pairs of high-resolution images for each scene instance, and each pair contains one noisy image and its corresponding clean image. Totally, these 320 pairs of images are used for training. For testing, we use the the SIDD validation set and the Darmstadt Noise Data set (DND) \cite{dnd}. DND does not provide any training data. It has 50 pairs of images captured by four different consumer cameras for testing. We got the PSNR and SSIM results by submitting the denoising images to the DND official website.

% \begin{figure}[ht]
% 	\centering{\includegraphics[width=0.9\columnwidth]{color_contrast.png}}
% 	\caption{Denoising results of the image from Kodak24 with noise level $\sigma=50$: (a) original image, (b) noisy image, (c)DnCNN\cite{dncnn}/25.80dB, (d) BRDNet\cite{brdnet}/26.33dB, (e) FFDNet\cite{ffdnet}/26.13dB, and (f) MHCNN/26.52dB.}
% 	\label{color_contrast}
% \end{figure}

\subsection{Experimental Results}
We evaluated the denoising performance of CDN on synthetic and real-world datasets and compared it with some popular methods.
\subsubsection{Synthetic noise}
Table~\ref{tabc} and Table~\ref{tabc2} shows the synthetic gray-scale noisy image denoising results of different methods on Set12 and BSD68, respectively. CDN outperforms other methods at all noise levels $\sigma=25$ and $\sigma=50$ on Set12 and $\sigma=15$ and $\sigma=25$ on BSD68. On set12, CDN has the largest improvement in denoising effect of image Barbara than other methods. As shown in Fig.~\ref{set12}, Barbara is rich in texture. CDN also performs well on other images with this property, such Monarch, Man and Couple.
Compared with other hierarchical networks U-Net \cite{UNet}, DIDN \cite{didn}, BRDNet \cite{brdnet} and NHNet \cite{NHNet}, CDN has superior performance. The visual comparison of denoised images is shown in Fig.~\ref{gray_contrast}. 
% The denoising results are shown in  Fig.~\ref{gray_contrast1}, where CDN shows the best visual effect. 

We also test CDN on the color-scale noisy image. As shown in Table \ref{tabcolor}, CDN achieves highest PSNR resutls on Kodak24 dataset. Fig.~\ref{color_contrast} and Fig.~\ref{color_contrast2} illustrate the visual results obtained from the CDN and other methods on Set5 and Kodak24, respectively. It can be observed that CDN can recover and obtain clearer images compared with the other methods.

\subsubsection{Real-world noise}
We used the SIDD validation set and DND to evaluate the performance of CDN on real-world image denoising. Table~\ref{real} lists the PSNR and SSIM results of different methods, where CDN achieves the best performance on the SIDD validation set and competitive performance on DND. Some denoised images of CDN on the SIDD dataset are shown in Fig.~\ref{figreal}, from which we can observe the noise has been removed successfully. 

\subsection{Ablation Experiments and Discussion} \label{ablation}

\subsubsection{Role of IIP}
IIP extracts the image information by image-to-image training. Therefore, it is valuable to confirm whether this information improves denoising results. 
We first carried out an ablation experiment by cutting off IIP in a trained CDN. This operation is accomplished by replacing the output of IIP with a zero matrix of the same size. Fig.~\ref{comp} shows the denoising results of this model, denoted as CDN-IIP. Compared with CDN, its denoised images are significantly more blurred, and the details are seriously missing. This result demonstrates that image information plays a key role in CDN restoring the image details.
We then study the PSNR and SSIM denoising results of removing IIP. Simply removing IIP definitely leads to a reduction of denoising performance, so we train CDN-IIP. CDN-IIP(R) in Table~\ref{contrast1} denotes removing IIP and retraining CDN. We observe that removing IIP will result in an overall decline in PSNR and SSIM, demonstrating image information's effectiveness. 

\subsubsection{Role of NEP}
NEP provides noise estimation of an noisy image. CDN-NEP in Fig.~\ref{comp} denotes CDN cutting off NEP but retaining IIP. It can be seen that although the denoised images of CDN-NEP still remain some image content information, the noise is obviously not removed well. Therefore, the noise distribution from NEP is essential for removing noise.
Similar to the study of IIP, we also report the PSNR and SSIM results of retained CDN-NEP. The results of CDN-NEP(R) on Set12 are listed in Table~\ref{contrast1}, which are significantly lower than that of CDN. It demonstrates that using the estimated noise distribution can improve denoising performance.

\subsubsection{Role of Training Methods}
CDN solves the limitations of traditional residual learning by using the image-to-image training method to train IIP and the similarity-based training method to train NEP. We study the effect of these training methods on the denoising performance. CDN-SSIM in Table~\ref{contrast1} denotes that training CDN without the image-to-image training method, which is implemented by removing the SSIM loss. Similarly, CDN-KLD denotes that training CDN without the similarity-based training method. CDN-SSIM-KLD denotes CDN without both the training methods, which can be considered as ordinary residual learning. The results in Table~\ref{contrast1} show CDN performs comprehensively better than that without the proposed training methods. CDN significantly outperforms CDN-SSIM-KLD, indicating that considering image information and self-similarity improves residual learning.

% There are two aspects of NEP that need to be explored. The first one is the contribution of NEP. The results of CDN without NEP on Set12 are listed in Table~\ref{contrast1}, which is significantly lower than that of CDN. It demonstrates that the estimited noise distribution improve denoising performance. The second one is the similarity-based training method. We study it by removing the KLD loss of NEP. Results in Table \ref{contrast1} shows CDN performs comprehensively better than that without KLD loss, demonstrating the effectiveness of the similarity-based training.  

% There are two aspects of NEP that need to be explored. The first one is the contribution of NEP. The results of CDN without NEP on Set12 are listed in Table~\ref{contrast1}, which is significantly lower than that of CDN. It demonstrates that the estimited noise distribution improve denoising performance. The second one is the similarity-based training method. We study it by removing the KLD loss of NEP. Results in Table \ref{contrast1} shows CDN performs comprehensively better than that without KLD loss, demonstrating the effectiveness of the similarity-based training.  

\section{Conclusion}
This paper suggests two limitations in residual learning for image denoising. It does not consider the image information and image self-similarity sufficiently. A novel denoising network CDN is proposed to solve the problems, achieving state-of-the-art results in synthetic and real-world image denoising. IIP in CDN is trained to extract image information with the image-to-image restoration method. Furthermore, a similarity-based training method is proposed to train NEP in CDN, which solves the second limitation. Ablation experiments demonstrate the effectiveness of the proposed methods. Previous studies in image denoising primarily focused on the network architecture. CDN provides a novel view that improves the training method for better denoising results, which will inspire future research.


\section*{Acknowledgment}
Supported by the National Natural Science Foundation of China(No. 11571325) and the Fundamental Research Funds for the Central Universities(No. CUC2019 A002).

\clearpage
\bibliography{IET-Submission-DoubleColumn-Template}

\begin{thebibliography}{10}

\bibitem{sidd}
Abdelrahman Abdelhamed, Stephen Lin, and Michael~S. Brown.
\newblock A high-quality denoising dataset for smartphone cameras.
\newblock In {\em 2018 IEEE/CVF Conference on Computer Vision and Pattern
  Recognition}, pages 1692--1700, 2018.

\bibitem{RID}
Saeed Anwar and Nick Barnes.
\newblock Real image denoising with feature attention.
\newblock In {\em 2019 IEEE/CVF International Conference on Computer Vision
  (ICCV)}, pages 3155--3164, 2019.

\bibitem{MCUNet}
Long Bao, Zengli Yang, Shuangquan Wang, Dongwoon Bai, and Jungwon Lee.
\newblock Real image denoising based on multi-scale residual dense block and
  cascaded u-net with block-connection.
\newblock In {\em Proceedings of the IEEE/CVF Conference on Computer Vision and
  Pattern Recognition Workshops}, pages 448--449, 2020.

\bibitem{nonlocal1}
A.~Buades, B.~Coll, and J.-M. Morel.
\newblock A non-local algorithm for image denoising.
\newblock In {\em 2005 IEEE Computer Society Conference on Computer Vision and
  Pattern Recognition (CVPR'05)}, volume~2, pages 60--65 vol. 2, 2005.


\bibitem{bm3d}
K.~{Dabov}, A.~{Foi}, V.~{Katkovnik}, and K.~{Egiazarian}.
\newblock Image denoising by sparse 3-d transform-domain collaborative
  filtering.
\newblock {\em IEEE Transactions on Image Processing}, 16(8):2080--2095, 2007.


\bibitem{NCSR}
Weisheng Dong, Lei Zhang, Guangming Shi, and Xin Li.
\newblock Nonlocally centralized sparse representation for image restoration.
\newblock {\em IEEE Transactions on Image Processing}, 22(4):1620--1630, 2013.

\bibitem{kodak24}
R.~Franzen.
\newblock Kodak lossless true color image suite: Photocd pcd0992.

\bibitem{wnnm}
S.~{Gu}, L.~{Zhang}, W.~{Zuo}, and X.~{Feng}.
\newblock Weighted nuclear norm minimization with application to image
  denoising.
\newblock In {\em 2014 IEEE Conference on Computer Vision and Pattern
  Recognition}, pages 2862--2869, 2014.

\bibitem{cbdnet}
Shi Guo, Zifei Yan, Kai Zhang, Wangmeng Zuo, and Lei Zhang.
\newblock Toward convolutional blind denoising of real photographs.
\newblock In {\em 2019 IEEE/CVF Conference on Computer Vision and Pattern
  Recognition (CVPR)}, pages 1712--1722, 2019.

\bibitem{prelu}
Kaiming He, Xiangyu Zhang, Shaoqing Ren, and Jian Sun.
\newblock Delving deep into rectifiers: Surpassing human-level performance on
  imagenet classification.
\newblock In {\em 2015 IEEE International Conference on Computer Vision
  (ICCV)}, pages 1026--1034, 2015.

\bibitem{resnet}
Kaiming He, Xiangyu Zhang, Shaoqing Ren, and Jian Sun.
\newblock Deep residual learning for image recognition.
\newblock In {\em Proceedings of the IEEE conference on computer vision and
  pattern recognition}, pages 770--778, 2016.

\bibitem{PSNR}
Alain Horé and Djemel Ziou.
\newblock Image quality metrics: Psnr vs. ssim.
\newblock In {\em 2010 20th International Conference on Pattern Recognition},
  pages 2366--2369, 2010.


\bibitem{bn}
Sergey Ioffe and Christian Szegedy.
\newblock Batch normalization: Accelerating deep network training by reducing
  internal covariate shift.
\newblock In {\em International conference on machine learning}, pages
  448--456. PMLR, 2015.


\bibitem{medical2}
Worku Jifara, Feng Jiang, Seungmin Rho, Maowei Cheng, and Shaohui Liu.
\newblock Medical image denoising using convolutional neural network: a
  residual learning approach.
\newblock {\em The Journal of Supercomputing}, 75(2):704--718, 2019.

\bibitem{kiku2013residual}
Daisuke Kiku, Yusuke Monno, Masayuki Tanaka, and Masatoshi Okutomi.
\newblock Residual interpolation for color image demosaicking.
\newblock In {\em 2013 IEEE international conference on image processing},
  pages 2304--2308. IEEE, 2013.

\bibitem{set5}
Jiwon Kim, Jung~Kwon Lee, and Kyoung~Mu Lee.
\newblock Accurate image super-resolution using very deep convolutional
  networks.
\newblock In {\em 2016 IEEE Conference on Computer Vision and Pattern
  Recognition (CVPR)}, pages 1646--1654, 2016.

\bibitem{adam}
Diederik~P Kingma and Jimmy Ba.
\newblock Adam: A method for stochastic optimization.
\newblock {\em arXiv preprint arXiv:1412.6980}, 2014.

\bibitem{relu}
Alex Krizhevsky, Ilya Sutskever, and Geoffrey~E Hinton.
\newblock Imagenet classification with deep convolutional neural networks.
\newblock {\em Advances in neural information processing systems},
  25:1097--1105, 2012.


\bibitem{udnet}
Stamatios Lefkimmiatis.
\newblock Universal denoising networks : A novel cnn architecture for image
  denoising.
\newblock In {\em 2018 IEEE/CVF Conference on Computer Vision and Pattern
  Recognition}, pages 3204--3213, 2018.

\bibitem{NLRN}
Ding Liu, Bihan Wen, Yuchen Fan, Chen~Change Loy, and Thomas~S. Huang.
\newblock Non-local recurrent network for image restoration.
\newblock In {\em Proceedings of the 32nd International Conference on Neural
  Information Processing Systems}, NIPS'18, page 1680–1689, Red Hook, NY,
  USA, 2018. Curran Associates Inc.

\bibitem{MWCNN}
P.~{Liu}, H.~{Zhang}, K.~{Zhang}, L.~{Lin}, and W.~{Zuo}.
\newblock Multi-level wavelet-cnn for image restoration.
\newblock In {\em 2018 IEEE/CVF Conference on Computer Vision and Pattern
  Recognition Workshops (CVPRW)}, pages 886--88609, 2018.

\bibitem{remote1}
Peng Liu, Meng Wang, Lizhe Wang, and Wei Han.
\newblock Remote-sensing image denoising with multi-sourced information.
\newblock {\em IEEE Journal of Selected Topics in Applied Earth Observations
  and Remote Sensing}, 12(2):660--674, 2019.


\bibitem{PANNet}
Ruijun Ma, Bob Zhang, Yicong Zhou, Zhengming Li, and Fangyuan Lei.
\newblock Pid controller-guided attention neural network learning for fast and
  effective real photographs denoising.
\newblock {\em IEEE Transactions on Neural Networks and Learning Systems},
  pages 1--14, 2021.

\bibitem{dhdn}
B.~{Park}, S.~{Yu}, and J.~{Jeong}.
\newblock Densely connected hierarchical network for image denoising.
\newblock In {\em 2019 IEEE/CVF Conference on Computer Vision and Pattern
  Recognition Workshops (CVPRW)}, pages 2104--2113, 2019.

\bibitem{DAVIS}
F.~Perazzi, J.~Pont-Tuset, B.~McWilliams, L.~{Van Gool}, M.~Gross, and
  A.~Sorkine-Hornung.
\newblock A benchmark dataset and evaluation methodology for video object
  segmentation.
\newblock In {\em Computer Vision and Pattern Recognition}, 2016.

\bibitem{dnd}
Tobias Plötz and Stefan Roth.
\newblock Benchmarking denoising algorithms with real photographs.
\newblock In {\em 2017 IEEE Conference on Computer Vision and Pattern
  Recognition (CVPR)}, pages 2750--2759, 2017.


\bibitem{DNResNet}
Haoyu Ren, Mostafa El{-}Khamy, and Jungwon Lee.
\newblock Dn-resnet: Efficient deep residual network for image denoising.
\newblock In C.~V. Jawahar, Hongdong Li, Greg Mori, and Konrad Schindler,
  editors, {\em Computer Vision - {ACCV} 2018 - 14th Asian Conference on
  Computer Vision, Perth, Australia, December 2-6, 2018, Revised Selected
  Papers, Part {V}}, volume 11365 of {\em Lecture Notes in Computer Science},
  pages 215--230. Springer, 2018.

\bibitem{UNet}
Olaf Ronneberger, Philipp Fischer, and Thomas Brox.
\newblock U-net: Convolutional networks for biomedical image segmentation.
\newblock In {\em International Conference on Medical image computing and
  computer-assisted intervention}, pages 234--241. Springer, 2015.

\bibitem{set12}
S.~Roth and M.J. Black.
\newblock Fields of experts: a framework for learning image priors.
\newblock In {\em 2005 IEEE Computer Society Conference on Computer Vision and
  Pattern Recognition (CVPR'05)}, volume~2, pages 860--867 vol. 2, 2005.

\bibitem{medical1}
Sameera V~Mohd Sagheer and Sudhish~N George.
\newblock A review on medical image denoising algorithms.
\newblock {\em Biomedical signal processing and control}, 61:102036, 2020.

\bibitem{pixelshuffle}
W.~Shi, J.~Caballero, F~Huszár, J.~Totz, and Z.~Wang.
\newblock Real-time single image and video super-resolution using an efficient
  sub-pixel convolutional neural network.
\newblock {\em 2016 IEEE Conference on Computer Vision and Pattern Recognition
  (CVPR)}, 2016.

\bibitem{resdnn}
G.~Singh, A.~Mittal, and N.~Aggarwal.
\newblock Resdnn: deep residual learning for natural image denoising.
\newblock {\em IET Image Processing}, 14(11):2425--2434, 2020.


\bibitem{adnet}
Chunwei Tian, Yong Xu, Zuoyong Li, Wangmeng Zuo, Lunke Fei, and Hong Liu.
\newblock Attention-guided cnn for image denoising.
\newblock {\em Neural Networks}, 124:117--129, 2020.

\bibitem{brdnet}
Chunwei Tian, Yong Xu, and Wangmeng Zuo.
\newblock Image denoising using deep cnn with batch renormalization.
\newblock {\em Neural Networks}, 121:461--473, 2020.

\bibitem{DudeNet}
Chunwei Tian, Yong Xu, Wangmeng Zuo, Bo~Du, Chia-Wen Lin, and David Zhang.
\newblock Designing and training of a dual cnn for image denoising.
\newblock {\em Knowledge-Based Systems}, 226:106949, 2021.

\bibitem{timofte2014a}
Radu Timofte, Vincent De~Smet, and Luc Van~Gool.
\newblock A+: Adjusted anchored neighborhood regression for fast
  super-resolution.
\newblock In {\em Asian conference on computer vision}, pages 111--126.
  Springer, 2014.


\bibitem{didn}
Songhyun Yu, Bumjun Park, and Jechang Jeong.
\newblock Deep iterative down-up cnn for image denoising.
\newblock In {\em Proceedings of the IEEE/CVF Conference on Computer Vision and
  Pattern Recognition Workshops}, pages 0--0, 2019.

\bibitem{vdn}
Zongsheng Yue, Hongwei Yong, Qian Zhao, Lei Zhang, and Deyu Meng.
\newblock Variational denoising network: Toward blind noise modeling and
  removal.
\newblock {\em arXiv preprint arXiv:1908.11314}, 2019.

\bibitem{2015Nonlocal}
Chenyang Zhang, Wenrui hu, Tianyu Jin, and Zhonglei Mei.
\newblock Nonlocal image denoising via adaptive tensor nuclear norm
  minimization.
\newblock {\em Neural Computing and Applications}, 29, 01 2018.

\bibitem{NHNet}
Jiahong Zhang, Lihong Cao, Tian Wang, Wenlong Fu, and Weiheng Shen.
\newblock Nhnet: A non-local hierarchical network for image denoising.
\newblock {\em IET Image Processing}, 2022.

\bibitem{mhcnn}
Jiahong Zhang, Meijun Qu, Ye~Wang, and Lihong Cao.
\newblock A multi-head convolutional neural network with multi-path attention
  improves image denoising.
\newblock {\em arXiv preprint arXiv:2204.12736}, 2022.

\bibitem{DRNet}
Jiahong Zhang, Yonggui Zhu, Wenyi Li, Wenlong Fu, and Lihong Cao.
\newblock Drnet: A deep neural network with multi-layer residual blocks
  improves image denoising.
\newblock {\em IEEE Access}, 9:79936--79946, 2021.

\bibitem{dncnn}
K.~{Zhang}, W.~{Zuo}, Y.~{Chen}, D.~{Meng}, and L.~{Zhang}.
\newblock Beyond a gaussian denoiser: Residual learning of deep cnn for image
  denoising.
\newblock {\em IEEE Transactions on Image Processing}, 26(7):3142--3155, 2017.

\bibitem{ffdnet}
Kai Zhang, Wangmeng Zuo, and Lei Zhang.
\newblock Ffdnet: Toward a fast and flexible solution for cnn-based image
  denoising.
\newblock {\em IEEE Transactions on Image Processing}, 27(9):4608--4622, 2018.


\bibitem{ssim}
Wang Zhou.
\newblock Image quality assessment: from error measurement to structural
  similarity.
\newblock {\em IEEE transactions on image processing}, 13:600--613, 2004.

\end{thebibliography}

\end{document}